\pdfoutput=1

\documentclass[11pt]{article}

\usepackage{ACL2023}

\usepackage{times}
\usepackage{latexsym}
\usepackage{booktabs}
\usepackage[T1]{fontenc}

\usepackage[utf8]{inputenc}

\usepackage{microtype}

\usepackage{inconsolata}
\usepackage{amsmath}

\title{xSIM++: An Improved Proxy to Bitext Mining Performance for Low-Resource Languages}

\author{Mingda Chen\thanks{~~Equal contribution}~, Kevin Heffernan$^*$, Onur Çelebi, Alex Mourachko, Holger Schwenk \\ 
  \texttt{\{mingdachen,kevinheffernan,celebio,alexmourachko,schwenk\}@meta.com} \\ Meta AI Research}

\usepackage{amssymb}
\usepackage{xspace}
\newcommand{\annote}[3]{{\color{#3}%
		\colorbox{#3}{\bfseries\sffamily\tiny\textcolor{white}{#2}}
		\color{#3}
		$\blacktriangleright$\footnotesize\emph{#1}$\blacktriangleleft$}%
}
\newcommand{\todo}[1]{\annote{#1}{TODO}{red}}
\newcommand{\mingda}[1]{\annote{#1}{Mingda}{violet}}
\newcommand{\holger}[1]{\annote{#1}{Holger}{blue}}

\renewcommand{\todo}[1]{}
\renewcommand{\mingda}[1]{}
\renewcommand{\holger}[1]{}
\renewcommand{\marginpar}[1]{}

\newcommand{\xsim}{\texttt{xsim}\xspace}
\newcommand{\xsimp}{\texttt{xsim++}\xspace}
\newcommand{\flores}{{\sc Flores200}\xspace}

\begin{document}
\maketitle

\begin{abstract}
We introduce a new proxy score for evaluating bitext mining based on  similarity in a multilingual embedding space: \xsimp. In comparison to \texttt{xsim}, this improved proxy leverages rule-based approaches to extend English sentences in any evaluation set with synthetic, hard-to-distinguish examples which more closely mirror the scenarios we encounter during large-scale mining. We validate this proxy by running a significant number of bitext mining experiments for a set of low-resource languages, and subsequently train NMT systems on the mined data. In comparison to \texttt{xsim}, we show that \xsimp is better correlated with the downstream BLEU scores of translation systems trained on mined bitexts, providing a reliable proxy of bitext mining performance without needing to run expensive bitext mining pipelines. \xsimp also reports performance for different error types, offering more fine-grained feedback for model development.\footnote{Code and data are available at \url{https://github.com/facebookresearch/LASER}.}
\end{abstract}

\section{Introduction}

When training neural machine translation (NMT) systems, it has been shown in prior works that generally, the quality of such systems increases with the availability of high-quality training data \citep{koehn2017six}. However, for many low-resource languages there are few public corpora available, posing many challenges. In order to address this sparsity, one approach is to supplement existing datasets with automatically created parallel corpora, and a technique which has shown to be successful for such issues is the task of bitext mining \citep{schwenk2019ccmatrix}.

In bitext mining, the aim is to find pairs of sentences with the same sentence meaning across collections of monolingual corpora. 
In this work, we adopt a \emph{global mining} approach \citep{schwenk2019wikimatrix}, which has shown recent success in providing high-quality data for low-resourced languages \citep{nllb}.

In order to evaluate any bitext mining method, a natural approach is to train a NMT system on the automatically created alignments. However, this is extremely costly. As an alternative, the BUCC task \citep{zweigenbaum2018overview} offers a method for evaluating bitext mining algorithms by embedding known alignments within monolingual corpora, and then reporting on the number of correctly aligned pairs. However, this task currently only covers 5 high-resourced languages (English, French, Russian, German and Chinese), and so is not applicable to the low-resource domain. In order to address this, another approach to evaluate bitext mining is to align existing multilingual parallel test sets. Two such test sets are Tatoeba\footnote{\url{https://github.com/facebookresearch/LASER/tree/main/data/tatoeba/v1}} and {\sc Flores200}.\footnote{\url{https://github.com/facebookresearch/flores/tree/main/flores200}} However, as shown by \citet{heffernan2022bitext}, the Tatoeba corpus is not very reliable given that for some sentence pairs there are only a few hundred sentences. Therefore, we opt to use {\sc Flores200}, which is also n-way parallel.

\begin{table*}[t]
    \centering\small
    \begin{tabular}{c|p{0.33\textwidth}|p{0.37\textwidth}}
        Transformation Category & \multicolumn{1}{c|}{Original Sentence} & \multicolumn{1}{c}{Transformed Sentence} \\ 
        \toprule
      Causality Alternation & Apart from the fever and a sore throat, I feel well and in \textcolor{red}{good} shape to carry out my work by telecommuting.  & Apart from the fever and a sore throat, I feel well and in \textcolor{red}{bad} shape to carry out my work by telecommuting \\
      \midrule
      Entity Replacement & \textcolor{red}{Charles} was the first member of \textcolor{red}{the British Royal Family} to be awarded a degree. & \textcolor{red}{M. Smith} was the first member of \textcolor{red}{The University} to be awarded a degree. \\
      \midrule
      Number Replacement & Nadal bagged \textcolor{red}{88\%} net points in the match winning \textcolor{red}{76} points in the \textcolor{red}{first} serve. & Nadal bagged \textcolor{red}{98\%} net points in the match winning \textcolor{red}{71} points in the \textcolor{red}{sixth} serve. \\
      \bottomrule
    \end{tabular}
    \caption{Examples of the transformations applied to the English sentences from \flores dev set. The red texts indicate the places of alternations.}
    \label{tab:transformation-examples}
\end{table*}

One existing method for evaluating bitext mining on parallel test sets is \xsim.\footnote{\url{https://github.com/facebookresearch/LASER/tree/main/tasks/xsim}} This method reports the error rate of misaligned sentences, and follows a margin-based global mining approach \citep{artetxe2018margin}. However, although using \xsim on test sets such as \flores has been shown to be useful as a proxy metric for bitext mining \citep{nllb}, it has the following limitations:
\begin{enumerate}
    \item Using \flores alone has proven to not be difficult enough as for many language pairs, existing approaches quickly saturate at 0\% error  \citep{nllb}.
    \item As the dev and devtest sets are quite small (997/1012 respectively), this is arguably not a good approximation for performance when mining against billions of possible candidate sentences.
    \item We have observed that there is not a significant overlap in the semantics between candidate sentences, meaning that it is not possible to test difficult scenarios that arise in bitext mining when choosing between multiple (similar) candidate pairs.
\end{enumerate}

\noindent In order to address these limitations, in this work we introduce \xsimp. This is an improved proxy for bitext mining performance which expands the dev and devtest sets of \flores to include both more data points, and also difficult to distinguish cases which provide far greater challenges to the models. Our contributions can be summarised as follows:
\begin{enumerate}
    \item We create a more semantically challenging and expanded English test set for \flores.
    \item We validate this new test set by independently performing 110 bitext mining runs, training 110 NMT systems on the output mined bitexts, and then determining both the correlation and statistical significance between \xsimp and the resulting BLEU scores.
    \item We open-source the expanded \flores dev and devtest sets, and also the \xsimp code to evaluate them\footnote{\url{https://github.com/facebookresearch/LASER}}.
\end{enumerate}

\section{Methodology}
\subsection{Background: \xsim}
Given two lists of sentences in different languages, \xsim seeks to align each sentence in the source language to a corresponding sentence in the target language based on a margin-based\footnote{In this work we report all results using the \emph{absolute} margin} similarity \citep{artetxe2018margin}. In doing so, \xsim leverages the mining approach described in \citet{artetxe-schwenk-2019-massively} to first encode sentences into embedding vectors, assign pairwise scores between sentences in the lists, and then take the sentence in the target language that achieves the maximum score as the final prediction. 
 \xsim relies on human-annotated parallel corpora and measures the performance of bitext mining using the fraction of misaligned source sentences, i.e., error rates.

\label{sec:xsim++}
\subsection{\xsimp}
As the effectiveness of \xsim is limited by the availability of parallel corpora, we choose to create \xsimp by automatically expanding the English sentences, and evaluate the sentence encoders on into-English language directions, following prior work on low-resource bitext mining \citep{heffernan2022bitext}.
Aside from the expanded candidate set, \xsimp follows the same procedure as \xsim.

\xsimp seeks to capture more subtle improvements in bitext mining by adding challenging negative examples. The examples are human-written sentences transformed by various operations. These operations intend to perturb semantics through minimal alternations in the surface text. In particular, we use the following categories of transformations: causality alternation, entity replacement, and number replacement. We focus on these three transformation types only as they easily allow us to create negative examples. Examples of the transformed sentences are shown in Table~\ref{tab:transformation-examples}. For these transformations, we adapt the implementation in \citet{dhole2021nlaugmenter}\footnote{Although this library has additional transformation methods available, many would create positive examples in this use case (e.g. paraphrases).} and describe the details of these transformations below.%

\paragraph{Causality Alternation.} To alter causality in a sentence, we (1) replace adjectives with their antonyms; (2) negate the meaning of sentences by adding or removing negation function words (e.g. ``did not'' and ``was not'') to the sentences; or (3) leverage the negation strengthening approach \citep{tan-etal-2021-causal}, which changes the causal relationships through more assertive function words (e.g. replacing ``may'' with ``will''). For example, as shown in \autoref{tab:transformation-examples} we replace ``good'' with the antonym ``bad''.

\paragraph{Entity Replacement.} We collect candidate entities from large amounts of monolingual data. Then we replace entities in sentences with the ones randomly sampled from the candidate set. For both stages, we use the named entity recognizer from {\sc nltk} \citep{bird2009natural}.

\paragraph{Number Replacement.} We use spaCy \cite{spacy2} to detect dates, ordinals, cardinals, times, numbers, and percentages and then randomly replace their values.\\

\begin{table}[t]
    \centering\small
    \begin{tabular}{l r r} 
                 & Total \# & \# per orig. \\
                 \toprule
      Original  &  997   &   -  \\
      \midrule
      Causality &  1868  &  1.87 \\
      Entity    &  37745 &  37.86 \\
      Number    &  3476  &  3.49 \\
      \bottomrule
    \end{tabular}
    \caption{Total numbers of original sentences and transformed sentences in different transformation categories. We also report the averaged numbers of transformations per original sentence for each category.}
    \label{tab:xsimplus-stats}
\end{table}

\noindent Given the strategies above, for each sentence we create multiple transformations (i.e. negative examples) of that source sentence. For example, consider \autoref{tab:transformation-examples}. In the ``Entity Replacement'' example we create a transformation by replacing two named entities. We can then continue this process by replacing these with other named entities until we have reached the desired number of total transformations\footnote{We set a maximum threshold of 100 transformations per category per sentence.}. Note that since the opportunity to change each category is dependent on the frequency of that category in the evaluation sets, some transformations occurred more than others (e.g. entities were more frequent than numbers). We summarize the data statistics for \xsimp on the \flores dev set in Table \ref{tab:xsimplus-stats}. Results for the devtest set are in appendix \ref{appendix-sec:flores-test-set}.

\section{Experiment}

In order to establish \xsimp as a proxy for bitext mining performance, we measure the correlation between both \xsim and \xsimp error rates, and the BLEU scores resulting from NMT systems trained on mined bitexts. More specifically, for each language we choose a sentence encoder model, followed by bitext mining using each respective encoder, and then train and evaluate bilingual NMT systems on the resulting mined bitexts. We use the \flores development sets when computing the BLEU scores. 

In order to validate \xsimp against varied embedding spaces, we encode (and mine) using two different multilingual encoder methods: LASER \citep{artetxe-schwenk-2019-massively} and LaBSE \citep{feng-etal-2022-language}. For LASER, we trained our own custom encoders (details below). For LaBSE, we used a publicly available model\footnote{https://huggingface.co/sentence-transformers/LaBSE} as the code and data for training LaBSE are not publicly available.

We randomly choose 10 low-resource languages to perform both encoder training (if applicable) and bitext mining. The languages are: Faroese (fao), Kabuverdianu (kea), Tok Pisin (tpi), Kikuyu (kik), Friulian (fur), Igbo (ibo), Luxembourgish (ltz), Swahili (swh), Zulu (zul), Bemba (bem). 
\holger{"randomly" from what pool? mention NLLB? Swahili is not really low-resource}

\paragraph{Encoder Training.}
We trained LASER encoders using the teacher-student approach described in \citet{heffernan2022bitext}. We choose a LASER model \citep{artetxe-schwenk-2019-massively} as our teacher, and then trained specialised students for each language. In order to train each student, we used both publicly available code\footnote{ \url{https://github.com/facebookresearch/fairseq/tree/nllb/examples/nllb/laser_distillation}} and bitexts (e.g. OPUS\footnote{\url{https://opus.nlpl.eu}}) \holger{mini-mine6 encoders also use BT data ...}

\paragraph{Bitext Mining.}
For each chosen encoder model, we perform bitext mining against approximately 3.7 billion sentences of English.
For low-resource languages, the sizes of monolingual data range from 140k to 124 million. Details are in the appendix.
We make use of monolingual data available from both Commoncrawl and Paracrawl\footnote{\url{https://paracrawl.eu}}, and operationalize the mining using the \texttt{stopes} library \cite{andrews-etal-2022-stopes}.\footnote{\url{https://github.com/facebookresearch/stopes}} For LASER, we use 1.06 as the margin threshold following \citet{heffernan2022bitext} and for LaBSE, we use 1.16.\footnote{We did grid search on threshold values from 1.11 to 1.25 on three languages (swh, ltz, and zul), decided the optimal one based on the BLEU scores, and used the threshold for the rest of languages.} Following mining, 
for each language we concatenate publicly available bitexts and the mined bitext as training data for NMT bilingual models using \texttt{fairseq},\footnote{\url{https://github.com/facebookresearch/fairseq}} translating from each foreign text into English. For all NMT systems, we keep the hyperparameters fixed (details in Appendix).

\paragraph{Evaluation.}
Model selection involves two use cases: comparisons within a model and across different models.
For the former comparison, given our custom encoders, we choose to compare 10 checkpoints from each model.\footnote{Evenly spaced between epochs 1 and 30.} For cross model comparisons, we compare each chosen encoder checkpoint against another existing system. In this case, the LaBSE encoder.  To quantitatively measure these two cases, we report pairwise ranking accuracy \cite{kocmi-etal-2021-ship} for \xsim and \xsimp. Formally, the accuracy is computed as follows

\begin{equation}
    \frac{\vert\text{s}(\text{proxy}\Delta)=\text{s}(\text{mining}\Delta)\text{ for all system pairs}\vert}{\vert\text{all system pairs}\vert}\nonumber
\end{equation}

\noindent where proxy$\Delta$ is the difference of the \xsim or \xsimp scores, mining$\Delta$ is the difference of the BLEU scores, $\text{s}(\cdot)$ is the sign function, and $\vert\cdot\vert$ returns the cardinal number of the input.

In this work, we have 550 system pairs with 55 pairs per language direction (i.e.  $11 \choose 2$ pairs given 10 custom LASER encoder checkpoints + LaBSE).
We always compare systems within a language direction as the scores for system pairs across different directions are not comparable.\footnote{There are factors varied across language directions that are unrelated to the quality of sentence encoders but could affect mining performance, such as amounts of monolingual data available for mining.}

\subsection{Results}

\begin{table}[t]
    \centering\small
    \begin{tabular}{l c c}
         Metric & Accuracy & GPU hours \\
         \toprule
       \xsim  & 35.48$\ $ & 0.43 \\
       \xsimp & 72.00$^*$ & 0.52 \\
       \midrule
       Mining BLEU (Oracle) & 100 & 19569 \\
       \bottomrule
    \end{tabular}
    \caption{
    Pairwise ranking accuracy along with the total number of GPU hours. For all experiments, we used NVIDIA A100 GPUs. An $*$ indicates that the result passes the significance test proposed by \citet{koehn-2004-statistical} with $p\text{-value} < 0.05$ when compared to \xsim.}
    \label{tab:main-result}
\end{table}

\begin{table}[t]
    \centering\small
    \begin{tabular}{l c}
         & Accuracy \\
         \toprule
       \xsimp & 72.00 \\
       \midrule
       Causality & 63.09 \\
       Entity & 65.45 \\
       Number & 60.73 \\
       Misaligned & 40.73 \\
       Causality + Entity & 68.55 \\
       Causality + Entity + Misaligned & 70.55 \\
       Causality + Misaligned & 68.00 \\
       Causality + Number & 66.73 \\
       Causality + Number + Misaligned & 71.45 \\
       Entity + Misaligned & 70.55 \\
       Number + Entity & 67.45 \\
       Number + Entity + Misaligned & 71.09 \\
       Number + Misaligned & 64.36 \\
       \bottomrule
    \end{tabular}
    \caption{Pairwise ranking accuracy when using combinations of error categories. Causality=Causality Alternation, Entity=Entity Replacement, Number=Number Replacement.}%
    \label{tab:analysis}
\end{table}

As shown in Table \ref{tab:main-result}, \xsimp significantly outperforms \xsim on the pairwise ranking accuracy. Additionally, when comparing the computational cost to mining, \xsimp costs over 99.9\% less GPU hours and saves approximately 3 metric tons of carbon emissions, but still manages to achieve a competitive accuracy. We observe similar trends for the within a model and across models use cases and report their separate accuracies in the appendix.

To better understand the contributions of each transformation category (cf. \autoref{sec:xsim++}) in measuring the final mining performance, we report accuracies for different combinations of categories in Table~\ref{tab:analysis}. In cases where an incorrect bitext alignment  do does not map to any of the augmented sentences of the true alignment, we denote these as ``misaligned''. We find that entity replacement helps most in improving the accuracy and combing all the transformations gives the best performance.

\section{Related Work}

As \xsimp uses rule-based data augmentation, it is related to work in other areas that also employ similar data augmentation methods, such as part-of-speech tagging \citep{sahin-steedman-2018-data}, contrastive learning \citep{tang-etal-2022-augcse}, text classification \citep{kobayashi-2018-contextual,wei-zou-2019-eda}, dialogue generation \citep{niu-bansal-2018-adversarial} and summarization \citep{chen-yang-2021-simple}.

\section{Conclusion and Future Work}

We proposed a proxy score \xsimp for bitext mining performance using three kinds of data augmentation techniques: causality alternation, entity replacement, and number replacement. To validate its effectiveness, we conducted large-scale bitext mining experiments for 10 low-resource languages, and reported pairwise ranking accuracies. We found that \xsimp significantly improves over \xsim, doubling the accuracies. Analysis reveals that entity replacement helps most in the improvement. In future work, we plan to extend \xsimp to non-English language pairs.

\section{Limitations}

We highlight three limitations of our work. The first is that \xsimp is automatically constructed. There could be noisy sentences leading to errors that are irrelevant to the quality of encoders. The second is that \xsimp applies transformations solely to English sentences. Generalizing it to non-English language pairs requires additional research. Finally, we have experimented with the two most popular multilingual encoders: LASER and LaBSE. There are other available approaches which would be interesting to also validate  \xsimp against.

\bibliography{xsim}
\bibliographystyle{acl_natbib}
\clearpage
\appendix
\section{Data Statistics for \xsimp with \flores devtest set}
\label{appendix-sec:flores-test-set}

\begin{table}[ht!]
    \centering\small
    \begin{tabular}{l r r} 
                 & Total \# & \# per orig. \\
                 \toprule
      Original  &  1012   &   -  \\ \midrule
      Causality &  1916  &  1.89 \\
      Entity    &  38855 &  38.39 \\
      Number    &  3262  &  3.22 \\ \bottomrule
    \end{tabular}
    \caption{Total numbers of original sentences and transformed sentences in different transformation categories. We also report the averaged numbers of transformations per original sentence for each category.}
    \label{tab:xsimplus-stats-appendix}
\end{table}

We report the data statistics for \xsimp with \flores devtest set in Table~\ref{tab:xsimplus-stats-appendix}.

\section{Sizes of Monolingual data for Low-Resource Languages}

\begin{table}[ht!]
    \centering\small
    \begin{tabular}{l r}
       Language  & Size \\
         \toprule
kik & 147,902 \\
kea & 226,507 \\
fur & 737,178 \\
fao & 1,179,475 \\
tpi & 1,661,743 \\
bem & 2,302,805 \\
ibo & 8,124,418 \\
zul & 20,477,331 \\
swh & 55,399,821 \\
ltz & 123,944,670 \\
       \bottomrule
    \end{tabular}
    \caption{Number of monolingual sentences for each language.}
    \label{tab:monolingual-data-size}
\end{table}

We report the sizes of monolingual data for each language in Table~\ref{tab:monolingual-data-size}.

\section{Hyperparameters for NMT systems}

\begin{table}[ht!]
    \centering\small
    \begin{tabular}{l r}
       \toprule
            encoder layers & 6 \\
            encoder attention heads & 8 \\
            encoder embed dim & 512 \\
            encoder FFNN embed dim & 4096 \\
            decoder layers & 6 \\
            decoder attention heads & 8 \\
            decoder embed dim & 512 \\
            decoder FFNN embed dim & 4096 \\
            optimiser & Adam \\
            adam betas & (0.9, 0.98) \\
            learning rate & 0.001 \\
            dropout & 0.3 \\
            spm vocab size & 7000 \\
       \bottomrule
    \end{tabular}
    \caption{Hyperparameters for NMT systems.}
    \label{tab:NMT-hyperparams}
\end{table}

We report hyperparameters for NMT evaluations in Table~\ref{tab:NMT-hyperparams}.

\section{Within and Across Model Accuracies}

\begin{table}[ht!]
    \centering\small
    \begin{tabular}{l c c}
         Metric & Within & Across \\
         \toprule
       \xsim  & 31.33$\ $ & 54.04$\ $ \\
       \xsimp & 69.77$^*$ & 82.00$^*$ \\
       \bottomrule
    \end{tabular}
    \caption{
    Pairwise ranking accuracy for comparisons within a model and across different models. An $*$ indicates that the result passes the significance test proposed by \citet{koehn-2004-statistical} with $p\text{-value} < 0.05$ when compared to \xsim.}
    \label{tab:within-across-result}
\end{table}

We report accuracies for within a model (i.e., LASER) and across different models (i.e., the 10 LASER checkpoints vs LaBSE) in Table~\ref{tab:within-across-result}. 

\end{document}